\documentclass[10pt,twocolumn,letterpaper]{article}

\usepackage{iccv}
\usepackage{times}
\usepackage{epsfig}
\usepackage{graphicx}
\usepackage{amsmath}
\usepackage{amssymb}
\usepackage{amssymb}% http://ctan.org/pkg/amssymb
\usepackage{pifont}% http://ctan.org/pkg/pifont
\newcommand{\cmark}{\ding{51}}%
\newcommand{\xmark}{\ding{55}}%
\usepackage[font=small,labelfont=bf,tableposition=top]{caption} % for the teaser
\usepackage{microtype}
\usepackage{color, colortbl}
\usepackage{comment}
\usepackage{float}
\usepackage{siunitx}
\usepackage{amsmath}
\usepackage[breaklinks=true,bookmarks=false]{hyperref} 
\hypersetup{
    linkcolor=black,
    urlcolor=red,
}

\DeclareMathOperator*{\argmin}{arg\,min}

\definecolor{FloorColor}{RGB}{189, 198, 255}
\newcommand{\objectCount}{3289 }
\newcommand{\objectCateogryCount}{187 }
\newcommand{\datasetScenes}{478 }
\newcommand{\datasetScans}{1482 }
\newcommand{\objectInstanceCount}{1947 }
\newcommand{\instanceCount}{48k }
\newcommand{\dbName}{3RScan} 

\newcommand{\OURS}{RIO}
\newcommand{\frameCount}{363k }

\newcommand\blfootnote[1]{%
	\begingroup
	\renewcommand\thefootnote{}\footnote{#1}%
	\addtocounter{footnote}{-1}%
	\endgroup
}

\iccvfinalcopy % *** Uncomment this line for the final submission
	
\def\iccvPaperID{1206} % *** Enter the ICCV Paper ID here

\ificcvfinal\fi
\begin{document}

%%%%%%%%% TITLE
\title{\OURS: 3D Object Instance Re-Localization in Changing Indoor Environments}

\author{
	Johanna Wald $^{1}$ \hspace{0.25cm} Armen Avetisyan $^{1}$ \hspace{0.25cm} Nassir Navab $^{1}$ \hspace{0.25cm}  Federico Tombari $^{1,2,*}$  \hspace{0.25cm} Matthias Nie{\ss}ner $^{1,*}$
	\vspace{0.1cm}
	\\
	$^{1}$ Technical University of Munich \hspace{0.3cm}
	$^{2}$ Google 
	\vspace{-0.1cm}
	\\
}

\twocolumn[{%val
	\renewcommand\twocolumn[1][]{#1}%
	\maketitle
	\begin{center}
		\vspace{-0.5cm}
		\captionsetup{type=figure}
   \includegraphics[width=1.0\linewidth]{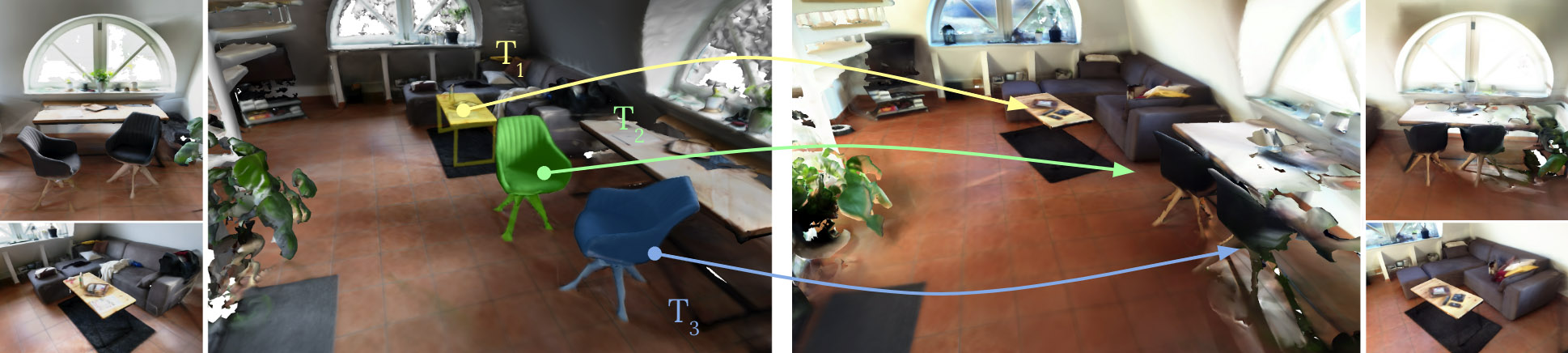}
		\captionof{figure}{
3D object instance re-localization benchmark: we want to robustly estimate the 6DoF pose ($\mathbf{T_1}, \mathbf{T_2}, ... \mathbf{T_n}$) of changed rigid object instances from a segmented source to a target scan taken at a later point in time.
    	}
    	\label{fig:teaser}
	\end{center}
}]

\begin{figure*}[t]
\begin{center}
   \includegraphics[width=1.0\linewidth]{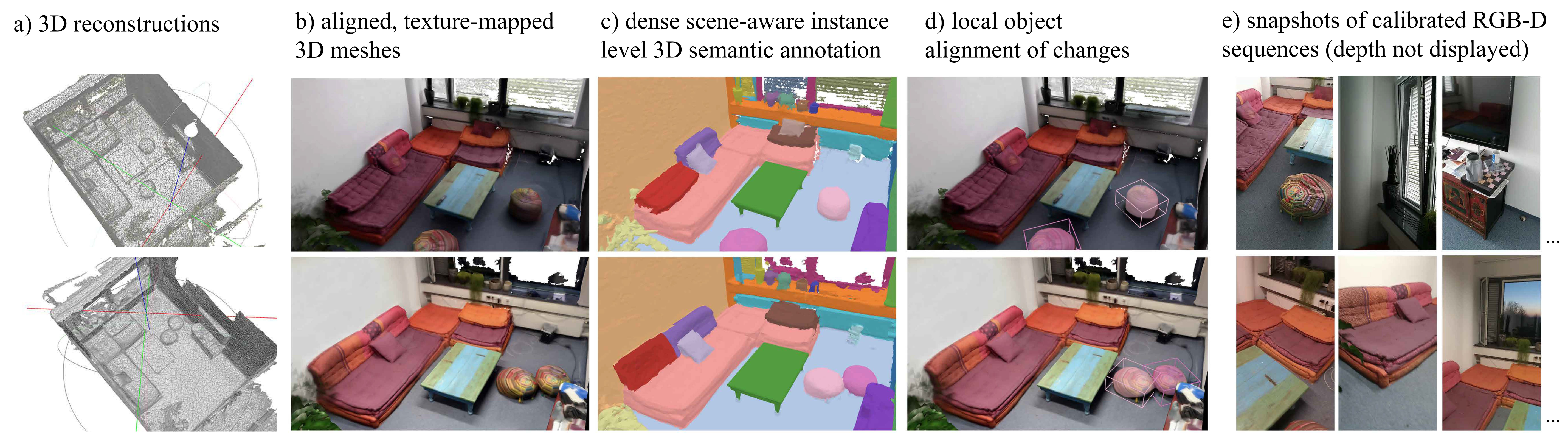}
\end{center}
   \caption{Example of a 3D scene pair of the \dbName{ } dataset. It provides: e) calibrated RGB-D sequences, a), b) aligned textured 3D reconstructions, c) dense instance-level semantic segmentation as well as d) symmetry-aware local alignment of changes.}
\label{}
\end{figure*}
\ificcvfinal\thispagestyle{empty}\fi

%%%%%%%%% ABSTRACT %
\begin{abstract}
In this work, we introduce the task of 3D object instance re-localization (\OURS): given one or multiple objects in an RGB-D scan, we want to estimate their corresponding 6DoF poses in another 3D scan of the same environment taken at a later point in time.
We consider \OURS{} a particularly important task in 3D vision since it enables a wide range of practical applications, including AI-assistants or robots that are asked to find a specific object in a 3D scene.
To address this problem, we first introduce \dbName, a novel dataset and benchmark, which features \datasetScans RGB-D scans of \datasetScenes environments across multiple time steps.
Each scene includes several objects whose positions change over time, together with ground truth annotations of object instances and their respective 6DoF mappings among re-scans.
Automatically finding 6DoF object poses leads to a particular challenging feature matching task due to varying partial observations and changes in the surrounding context.
To this end, we introduce a new data-driven approach that efficiently finds matching features using a fully-convolutional 3D correspondence network operating on multiple spatial scales. Combined with a 6DoF pose optimization, our method outperforms state-of-the-art baselines on our newly-established benchmark, achieving an accuracy of 30.58\%.
\end{abstract}

\blfootnote{* Authors share senior authorship.}

%%%%%%%%% BODY TEXT
\section{Introduction}
3D scanning and understanding of indoor environments is a fundamental research direction in computer vision laying the foundation for a large variety of applications ranging from indoor robotics to augmented and virtual reality. 
In particular, the rapid progress in RGB-D scanning systems \cite{Newcombe2011,Niessner2013,Whelan2015,Dai2017bundlefusion}
allows to obtain 3D reconstructions of indoor scenes using only low-cost scanning devices such as the Microsoft Kinect, Intel Real Sense, or Google Tango.
Along with the ability to capture 3D maps, researchers have shown significant interest in using these representations to perform 3D scene understanding and developed a rapidly-emerging line of research focusing on tasks such as 3D semantic segmentation \cite{Dai2017,Ruizhongtai2016,Song2016} or 3D instance segmentation \cite{Hou2018}. 
However, the shared commonality between these works is that they only consider static scene environments. 
In this work, we focus on environments that change over time.
Specifically, we introduce the task of object instance re-localization (\OURS): given one or multiple objects in an RGB-D scan, we want to estimate their corresponding 6DoF poses in another 3D scan of the same environment taken at a different point in time. Therefore, the captured reconstructions naturally cover a variety of temporal changes; see Fig.~\ref{fig:teaser}. 
We believe this is a critical task for many indoor applications, for instance, for a robot or virtual assistant to find a specific object in its surrounding environment.
\blfootnote{{\tt\href{https://waldjohannau.github.io/RIO/}{https://waldjohannau.github.io/RIO/}}}

The main challenge in \OURS \, -- finding the 6DoF of each object -- lies in establishing good correspondences between re-scans, which is non-trivial due to different scanning patterns and changing geometric context. 
These make the use of hand-crafted geometric descriptors, such as FPFH \cite{Rusu2009} or SHOT \cite{Tombari2010}, less effective. 
Similarly, learned 3D feature matching approaches, such as 3DMatch~\cite{Zeng2016,Deng2018}, cannot be easily leveraged since they are trained on self-supervised correspondences from static 3D scenes, and are hence very susceptible to geometry changes. 
One of the major limitations in using data-driven approaches for object instance localization is the scarce availability of supervised training data. While existing RGB-D datasets, such as ScanNet \cite{Dai2017} or SUN RGB-D \cite{Song2015}, provide semantic segmentations for hundreds of scenes, they lack temporal annotations across scene changes. In order to address this shortcoming, we introduce \dbName{} a new dataset that is composed of \datasetScans RGB-D sequences. 
An essential novelty of the proposed dataset is that several re-scans are provided for every environment. 
The dataset includes not only dense ground truth semantic instance annotations (for every scan), but also associates objects that have changed in appearance and/or location between re-scans. 
In addition to using \dbName{} for training feature descriptors, we also introduce a new benchmark for object instance localization.

In order to learn from this data, we propose a fully-convolutional multi-scale network capable of learning geometric features in dynamic environments. 
The network is trained with corresponding TSDF (truncated signed distance function) patches on moved objects extracted at two different spatial scales in a self-supervised fashion. 
As a result, we obtain change-invariant local features that outperform state-of-the-art baselines in correspondence matching and on our newly created benchmark for re-localization of object instances. 
In summary, we explore the task of 3D Object Instance Re-Localization in changing environments and contribute:
\begin{itemize}
    \item \dbName, a large indoor RGB-D dataset of changing environments that are scanned multiple times. We provide ground truth annotations for dense semantic instance labels and changed object associations.
    \item a new data-driven object instance re-localization approach that learns robust features in changing 3D contexts based on a geometric multi-scale neural network.
\end{itemize}

%-------------------------------------------------------------------------
\section{Related Work}

\begin{table*}[t]
\centering
\caption{RGB-D indoor datasets for 3D scene understanding: We list synthetic as well as real dataset and compare their size together with other properties such as the availability of scene changes.}
\label{table:datasets}
\resizebox{\textwidth}{!}{%
{\renewcommand{\arraystretch}{1.3}%
\begin{tabular}{lcclccc}
\begin{tabular}[c]{@{}c@{}}\textbf{Dataset}\end{tabular} & 
\begin{tabular}[c]{@{}c@{}}\textbf{Size}\end{tabular} & 
\begin{tabular}[c]{@{}c@{}}\textbf{Real}\end{tabular} &
\begin{tabular}[c]{@{}c@{}}\textbf{Data Acquisition / Generation}\end{tabular} &
\begin{tabular}[c]{@{}c@{}}\textbf{Benchmarks}\end{tabular} &
\begin{tabular}[c]{@{}c@{}}\textbf{Changes}\end{tabular}\\ 
\hline
NYUv2 \cite{Silberman2012} & 464 scenes & \cmark & recordings with Kinect & Depth and Semantics & \xmark \\
SUN RGB-D \cite{Song2015} & 10k frames & \cmark & recordings with 4 different sensors & 3D Object Detection & \xmark \\       
SUN-CG \cite{Song2016} & 45K rooms, 500K images & \xmark & rendered, layout hand-designed &  Scene-Completion & \xmark \\
ScanNet \cite{Dai2017} & 1513 scans, 2.5M images & \cmark & recordings with Structure Sensor & Semantic Voxel Labeling & \xmark \\
Fehr et al. \cite{Fehr2017} & 23 scans of 3 scenes & \cmark & recordings with Tango & Change Detection & \cmark \\
Matterport3D \cite{Chang2017} & 90 buildings, $\sim$ 200k images & \cmark & recordings with Matterport & several & \xmark \\
SceneNet RGB-D \cite{McCormac2017} & 15K trajectories, 5M images & \xmark & photo-realistic, random scenes & SLAM & \xmark \\
InteriorNet \cite{Li2018} & \textit{millions} / unknown & \xmark & photo-realistic, layout hand-designed & SLAM & \cmark \\
RGB Reloc \cite{Valentin2016} & 4 scenes, 12 rooms & \cmark & recording with Kinect & Camera Re-Localization & \xmark \\
\textbf{\dbName{ } (Ours)} & \datasetScans scans of \datasetScenes scenes & \cmark & recordings with Tango & Object Instance Re-Localization & \cmark \\
\end{tabular}}}
\end{table*}

\paragraph{3D Object Localization and Keypoint Matching} 
3D object localization and pose estimation via keypoint matching are long standing areas of interest in computer vision. 
Until very recently, 3D hand-crafted descriptors \cite{Tombari2010,Rusu2009} where prominently used to localize objects under occlusion and clutter by determining 3D point-to-point correspondences. 
However with the success of machine learning, the interest shifted to deep learned 3D feature descriptors capable of embedding 3D data, such as meshes or point clouds, in a discriminative latent space \cite{Zeng2016,Ruizhongtai2016,Deng2018}. 
Even though these approaches show impressive results on tasks such as correspondences matching and registration, they are restricted to static environments.
In this work, we go one step further by focusing on dynamic tasks; specifically, we aim to localize given 3D objects from a source scan in a cluttered target scan which contains common geometric and appearance changes.

\paragraph{RGB-D Scene Understanding}
Scene understanding methods based on RGB-D data generally rely on volumetric or surfel-based SLAM to reconstruct the 3D geometry of the scene while fusing semantic segments extracted via Random Forests \cite{Valentin2015,Wald2018} or CNNs \cite{McCormac2016, Nakajima2018}. 
Other works such as SLAM++ \cite{Renato2013} or Fusion++ \cite{McCormac2018} operate on an object level and create semantic scene graphs for SLAM and loop closure. 
Non-incremental scene understanding methods, in contrast, process a 3D scan directly to obtain semantic, instance or part segmentation \cite{Ruizhongtai2016,Ruizhongtai2017,Rethage2018, Dai2018,Hou2018}. 
Independently from the approach, all these methods rely on the assumption that objects are static and the scene structure does not change over time.

\paragraph{RGB-D Datasets}
Driven by the great interest in the development of scene understanding applications, several large-scale datasets based on RGB-D data have been recently proposed \cite{Firman2016}. 
We have summarized the most prominent efforts in Table~\ref{table:datasets}, together with their main features (\textit{e.g.}, number of scenes, mean of acquisition). 
The majority of datasets do not include changes in the scene layout and objects therein, and assume each scene is static over time. 
This is the case of ScanNet \cite{Dai2017}, currently the largest real dataset for indoor scene understanding consisting of $1500$ scans of approx. $750$ unique scenes. 
Notably, only a few recent proposals started exploring the idea of collecting scene changes to allow long-term scene understanding. InteriorNet \cite{Li2018} is a large-scale synthetic dataset, in which random physics-based furniture shuffles and illumination changes are applied to generate appearance and geometry variations which indoor scenes typically undergo. 
Several state-of-the-art sparse and dense SLAM approaches are compared on this benchmark. 
Despite the impressive size and indisputable usefulness, we argue that, due to the domain gap between real and synthetic imagery, the availability of real sequences remain crucial for the development of long-term scene understanding. 
To the best of our knowledge, the only real dataset encompassing scene changes is the one released by Fehr \etal\cite{Fehr2017}, which includes 23 sequences of 3 different rooms used to segment the scene structure from the movable furniture, though lacking the annotations and necessary size to train and test current learned approaches. 
\section{\dbName-Dataset}

We propose \textbf{3R}Scan, a large scale, \textbf{R}eal-world dataset which contains multiple ($2-12$) 3D snapshots (\textbf{R}e-scans) of naturally changing indoor environments, designed for benchmarking emerging tasks such as long-term SLAM, scene change detection \cite{Fehr2017} and camera or object instance \textbf{R}e-Localization.
In this section, we describe the acquisition of the scene scans under dynamic layout and moving objects, as well as that of annotation in terms of object pose and semantic segmentation. 

\subsection{Overview}
\begin{figure*}[htbp]
\begin{center}
   \includegraphics[width=1.0\linewidth]{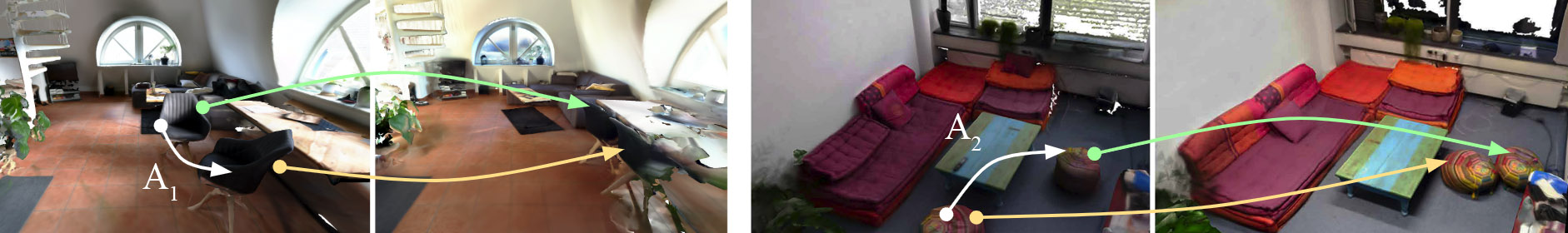}
\end{center}
   \caption{Instance ambiguities in presence of scene changes: since the instance mapping is unknown, multiple solutions are plausible, which we provide in our dataset from user annotations indicating all possibilities.}
\label{fig:ambiguity}
\end{figure*}

The recorded sequences are either a) controlled, where pairs are acquired within a time frame of only a few minutes under known scene changes or b) uncontrolled, where unknown changes naturally occurred over time (up to a few months) via scene-user interaction. 
All \datasetScans sequences were recorded with a Tango mobile application to enable easy usage for untrained users. 
Each sequence was processed offline to get bundle-adjusted camera poses with loop-closure and texture mapped 3D reconstructions. 
To ensure high variability, $45+$ different people recorded data in more than $13$ different countries. 
Each sequence comes with aligned semantically annotated 3D data and corresponding 2D frames (approximately \frameCount in total), containing in detail:
\begin{itemize}
\item calibrated RGB-D sequences with variable $n$ RGB $\mathbf{R_i}, ... \mathbf{R_n}$ and depth images $\mathbf{D_i}, ... \mathbf{D_n}$. 
\item textured 3D meshes 
\item camera poses $\mathbf{P_i}, ... \mathbf{P_n}$ and calibration parameters $\mathbf{K}$. 
\item global alignment among scans from the same scene as a global transformation $\mathbf{T}$.
\item dense instance-level semantic segmentation where each instance has a fixed ID that is kept consistent across different sequences of the same environment.
\item object alignment, \ie a ground truth transformation $\mathbf{T_{GT}} = \mathbf{R_{GT}} + \mathbf{t_{GT}}$ for each changed object together with its symmetry property. %
\item intra-class transformations $\mathbf{A}$ of ambiguous instances in the reference to recover all valid object poses in the re-scans (see Figure \ref{fig:ambiguity}).
\end{itemize}

\subsection{Scene Changes}
 
Due to the repetitive recording of interactive indoor environments, 
our data naturally captures a large variety of temporal scene changes.
Those changes are mostly rigid and include a) objects being moved (from a few centimeters up to a few meters) or b) objects being removed or added to the scene. 
Additionally, non-rigid objects such as curtains or blankets and the presence of lighting changes create additional challenging scenarios. 

\subsection{Annotation}

The dataset comes with rich annotations which include scan-to-scene-mappings and 3D transformations (section \ref{sec:instance_alignment}) together with dense instance segmentation (section \ref{semantic_segmentation}). 
More details and statistics regarding the annotations are given in the supplementary material.

\subsubsection{Semantic Segmentation}
\label{semantic_segmentation}

\begin{figure}[b]
\begin{center}
   \includegraphics[width=1.0\linewidth]{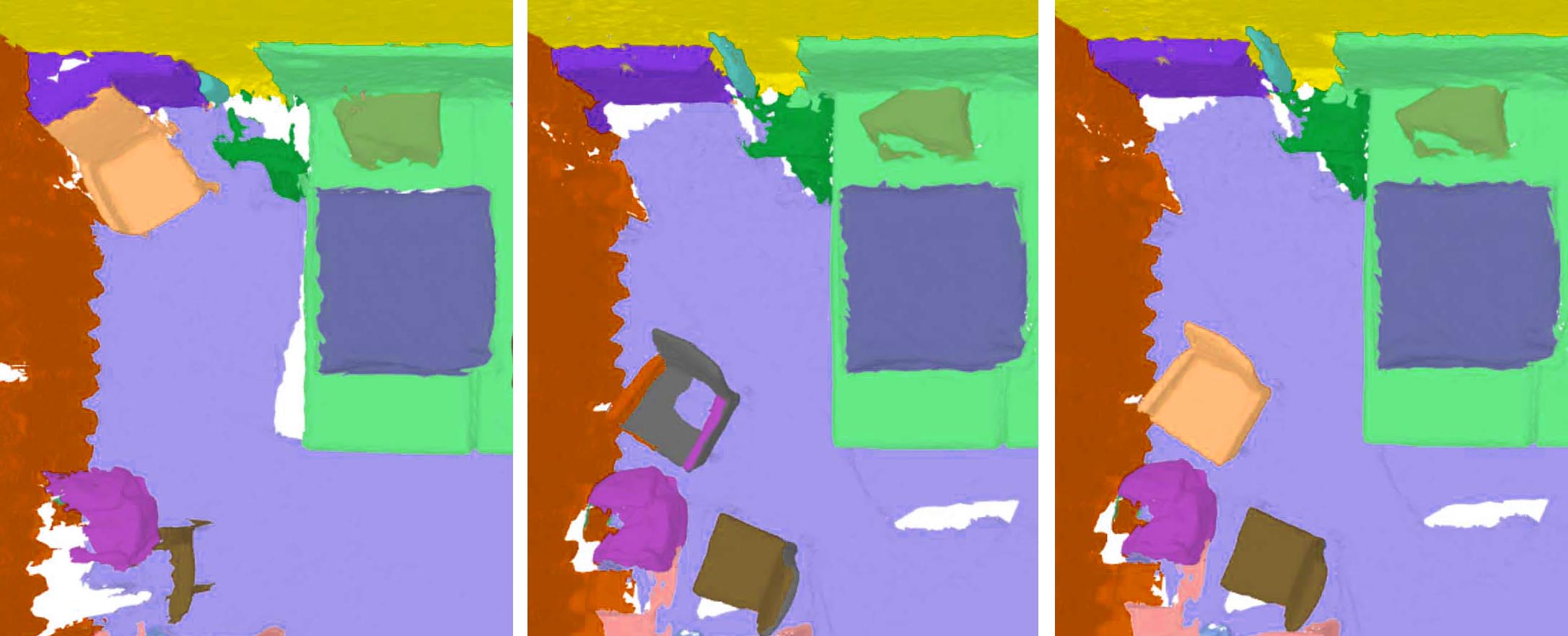}
\end{center}
   \caption{Propagation result from a reference (left) to a re-scan (center) and the manual cleanup (right). Please note the false propagations in the presence of scenes changes: here the orange \textit{armchair} was moved and its label was therefore incorrectly propagated.}
\label{fig:propagation}
\end{figure}

Similarly to ScanNet~\cite{Dai2017}, instance-level semantic annotations are obtained by labeling on a segmented 3D surface directly. For this, each reference scan was annotated with a modified version of ScanNet's publicly available annotation framework. To reduce annotation time, we propagate the annotations in a segment-based fashion from the reference scan to each re-scan using the global alignment $\mathbf{T}$ with the scan-to-scene mappings. This gives us very good annotation estimates for the re-scans, with the assumption that most parts of the scene remain static. Figure \ref{fig:propagation} gives an example of automatic label propagation from a hand-annotated scene in the presence of noise and scene changes. Semantic segments were annotated by human experts using a web-based crowd-sourcing interface and verified by the authors. The average annotation coverage of the semantic segmentation for the entire dataset is $98.5\%$.

\subsubsection{Instance Changes}
\label{sec:instance_alignment}
\begin{figure}[htbp]
\begin{center}
   \includegraphics[width=1.0\linewidth]{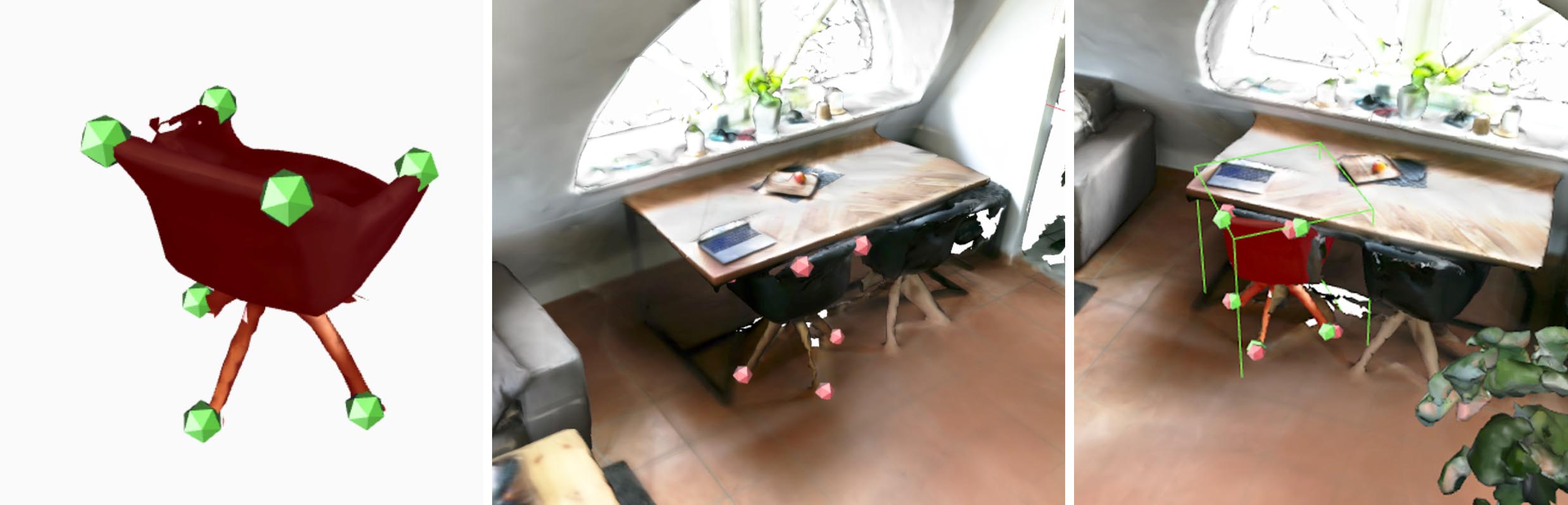}
\end{center}
   \caption{Example of the correspondences-based 3D instance alignment (right). 3D transformations are computed by manual annotation of corresponding keypoints on the objects (left, green) and the scene (center, red) respectively.}
\label{fig:alignment_tool}
\end{figure}

To obtain instance-level 3D transformations, a keypoint-based 3D annotation and verification interface was developed based on the CAD alignment tool used in \cite{Avetisyan2018}. A 3D transformation is obtained by applying Procrustes on manually annotated 3D keypoint correspondences on the object from the reference and its counterpart in the re-scan (see Figure \ref{fig:alignment_tool}). Additionally to this 3D transformation, a symmetry property was assigned to each instance. 
\subsection{Benchmark}

Based on this data, we set up a public benchmark for 3D instance-level object re-localization in changing indoor environments. Given one or multiple objects in a segmented source scene, we want to estimate the corresponding 6DoF poses in a target scan of the same environment taken at a different point in time. Namely, transformations $\mathbf{T_1} = \mathbf{R_1} + \mathbf{t_1}, ..., \mathbf{T_m}$ as a translation $\mathbf{t_1}, .. \mathbf{t_m}$ and rotation $\mathbf{R_1}, ..., \mathbf{R_m}$ need to be detected for all given $m$ instances in A (left Figure \ref{fig:teaser}) to instances in B (right). Predictions are evaluated against the annotated 3D transformation. A 6DoF pose estimation is considered successful if the translation and rotation error to the given ground truth transformation is within a small range. In our experiments we set these thresholds to $t \leq \SI{10}{\cm}$ and $r \leq \SI{10}{\degree}$ and $t \leq \SI{20}{\cm}$ and $r \leq \SI{20}{\degree}$ respectively. To avoid misalignment of symmetric objects, the respective symmetry property is considered. 
We publicly release our dataset with a standardized test, validation and training set (see Table \ref{table:data_split}) and all mentioned annotations. To allow a fair comparison of different methods, we also release a hidden test set together with an automatic server-side testing script.

\begin{table}[htbp]
\centering
\caption{Statistics on the test, train and validation set of \dbName.}
\label{table:data_split}
\begin{tabular}{l|llll}
 & test & train & validation & total  \\
\hline
\#scenes & 46 & 385 & 47 & \textbf{478} \\
\#re-scans & 101 & 793 & 110 & \textbf{1004} \\
\#scans & 147 & 1178 & 157 & \textbf{1482}\\
\end{tabular}
\end{table}

\section{3D Object Instance Re-Localization}

In order to address the task of \OURS, we propose a new data-driven approach that finds matching features in changing 3D scans using a 3D correspondence network. 
Our network operates on multiple spatial scales to encode change-invariant neighborhood information around the object and the scene. 
Object instances are re-localized by combining the learned correspondences with RANSAC and a 6DoF object pose optimization.

\begin{figure*}[t]
\begin{center}
   \includegraphics[width=1.0\linewidth]{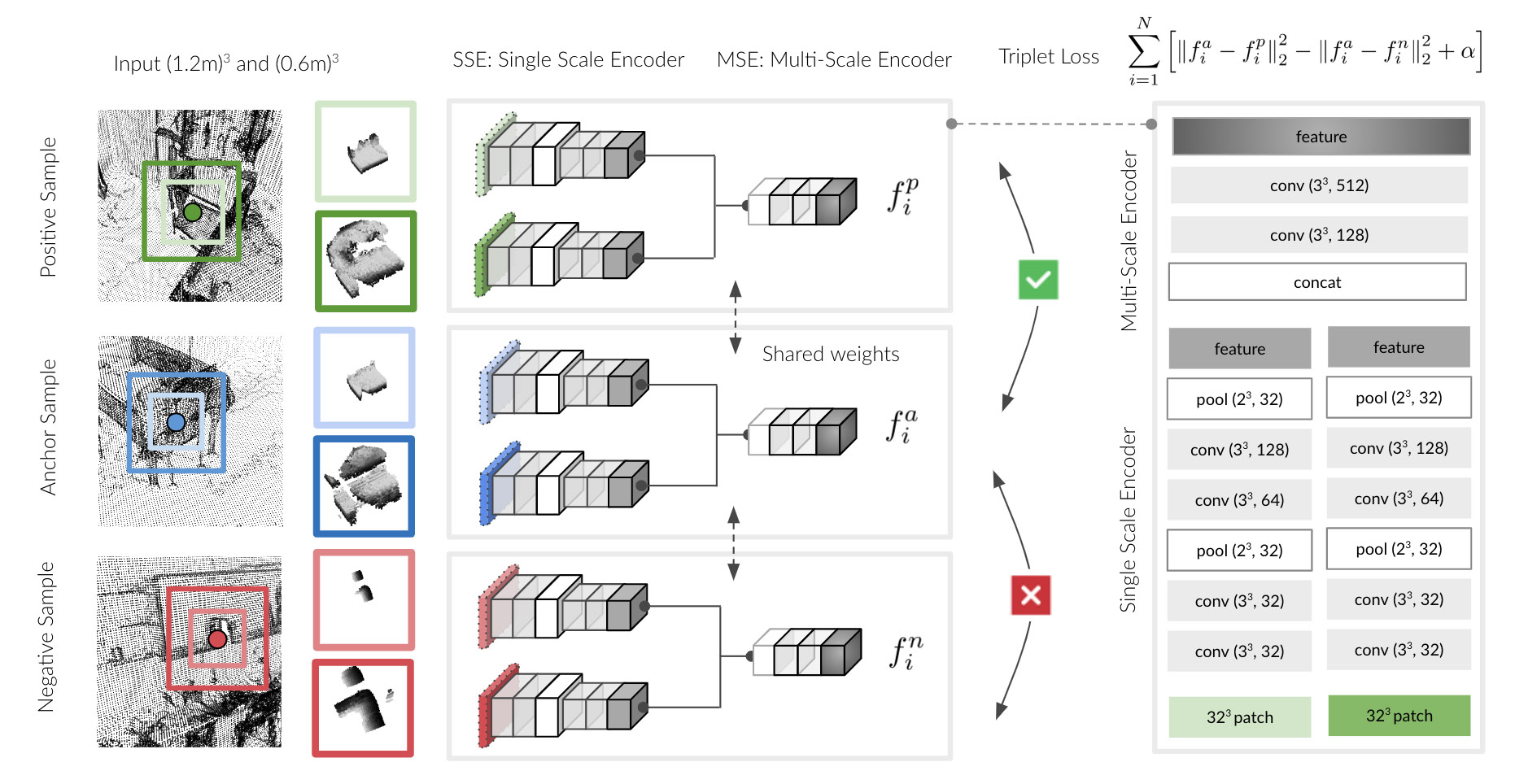}
\end{center}
   \caption{Our multi-scale triplet network architecture: during training, each anchor (blue) is paired with a positive (green) and a negative sample (red). The network minimizes the distance between the positive samples and maximizes negative sample distances by processing two scales in a separate branch each. Weights are shared in each SSE block of the same size and the MSE.}
\label{fig:network}
\end{figure*}

% methodology

\subsection{Data Representation}

The input of our network are TSDF patches. For each 3D keypoint on the source object or the target scene, the surrounding 3D volume is extracted at two different scales. They are chosen to be $32 \times 32 \times 32$ voxel grids containing TSDF values with spatial resolutions of $(\SI{1.2}{\m})^3$ and $(\SI{0.6}{\m})^3$. Their corresponding voxel sizes are \SI{1.875}{\cm} and \SI{3.75}{\cm}. 

\subsection{Network Architecture}

The network architecture of \OURS{ } is visualized in Figure~\ref{fig:network}. Due to non-padded convolutions and two pooling layers the input volumes are reduced to a 512-dimensional feature vector. It consists of two separate single scale encoders (SSE) and a subsequent multi-scale encoder (MSE). The two different input resolutions capture different neighborhoods with a different level of detail. Since both single scale encoder branches are identical, their network responses are concatenated before being fed into the MSE, as visualized in Figure~\ref{fig:network}. This multi-scale architecture helps to simultaneously capture fine geometric details as well as higher-level semantics of the surroundings. We show that our multi-resolution network produces richer features and therefore outperforms single scale architectures that process each scale independently by a large margin. Please also note that the two network branches do not share weights since they process the geometry of different context. To achieve a strong gradient near the object surface the raw TSDF is inverted in the first layer of the network such that
\begin{equation}
    \hat{TSDF} = 1 - \left| TSDF \right|.
\end{equation}

\subsection{Training}

During training, a triplet network architecture together with a triplet loss (equation \ref{eq:tripletloss}) is used. It maximizes the $L_2$ distance of negative patches and minimizes the $L_2$ distance of positive patches. We choose the margin $\alpha$ to be $1$. For optimization, Adam optimizer with an initial learning rate of $0.001$ is used.

\begin{equation}
\label{eq:tripletloss}
    \sum_{i=1}^N{N} \Big[||f_i^{a} - f_i^{p}||_2^2 - ||f_i^{a} - f_i^{n}||_2^2 + \alpha\Big]
\end{equation}

\subsection{Training Data: From Static to Dynamic}

\begin{figure*}[ht!]
    \centering
    \includegraphics[width=0.95\linewidth]{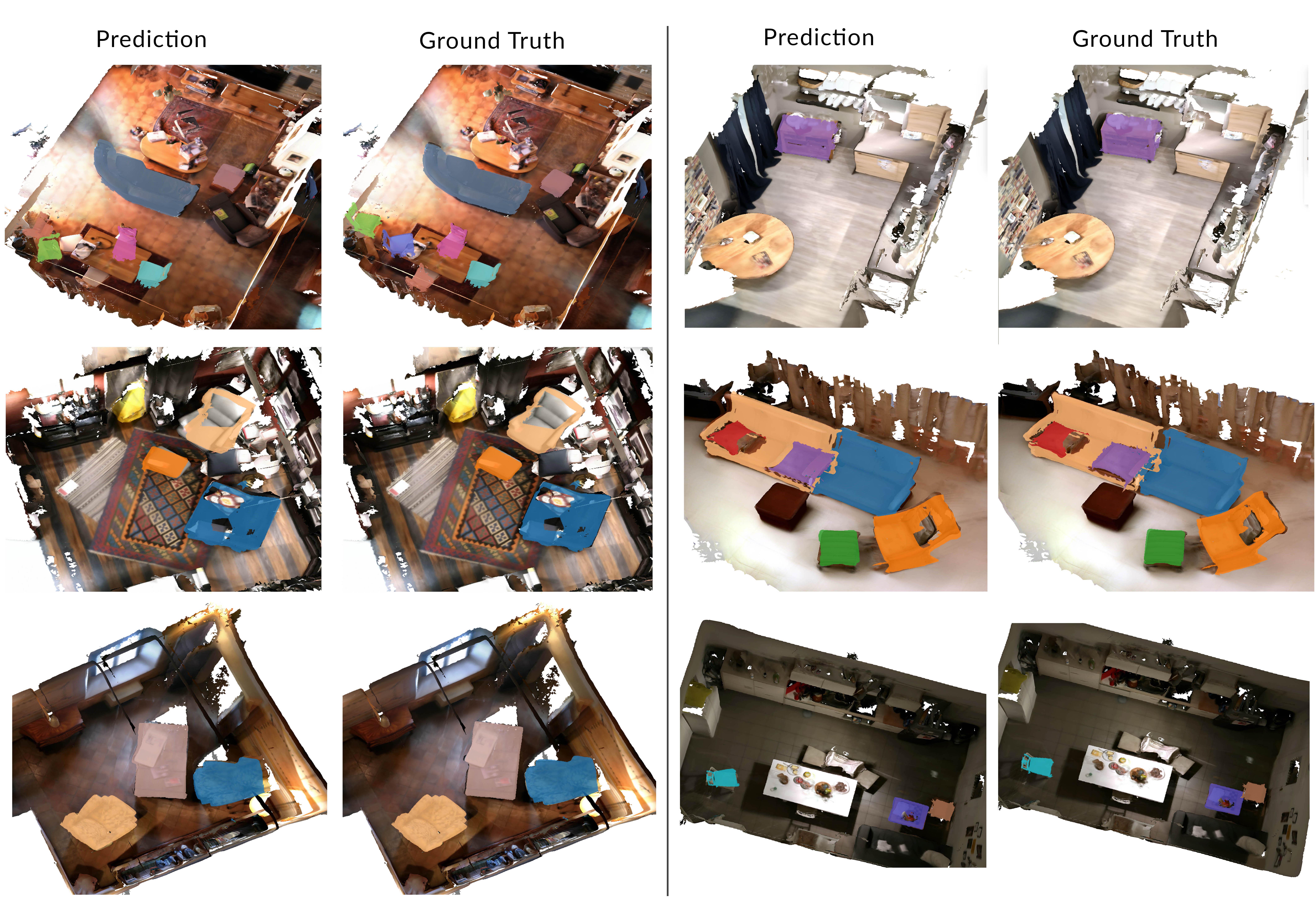}
    \caption{Qualitative results of 3D rigid object instance re-localization (\OURS) of our learned multi-scale method in different changing environments. Different instances, taken from the reference scan, are visualized with different colors on top of the re-scan.}
    \label{fig:results}
\end{figure*}

We initially train our network fully self-supervised with static TSDF patches extracted from RGB-D sequences of our dataset. To be able to deal with partial reconstructions induced by different scanning patterns, two small sets of non-overlapping frames are processed to produce two different TSDF volumes of the same scene. Then, first Harris 3D keypoints are extracted on one volume, then these same locations are refined on the other volume via non-maxima suppression of the Harris responses within a small radius around each extracted keypoint.  
If corresponding keypoints on the two volumes are above a certain threshold, we consider them a suitable patch pair and use it for pre-training of our network.

The goal of our method is to produce a local feature encoding that maps the local neighborhood of an object around a 3D keypoint on a 3D surface to a vector while being invariant to local changes around the object of interest. We learn this change-invariant descriptor by using the object alignments and sampling dynamic patches from our proposed \dbName-dataset. So, once converged we fine-tune the static network with dynamic 3D patches specifically generated around points of interest on moving objects. To learn only higher level features,  during fine-tuning, we freeze the first layers and only train the multi-scale encoder branch of our network. Correspondence pairs are generated in a self-supervised fashion while using the ground truth pose annotations of our training set to find high keypoint responses in the same small radius around each source 3D keypoint. The negative counterpart of each triplet is randomly selected from another training scene but also includes TSDF patches on removed objects. Random rotation augmentation is applied to enlarge our training data.

\vspace{0.3cm}

\subsection{6DoF Pose Alignment}

To re-localize object instances, we first compute features for keypoints on the source objects and the whole target scene. Correspondences for the model keypoints are then found via k-nearest neighbour search in the latent space of the feature encoding of the points in the scene. After outliers are filtered with RANSAC, remaining correspondences serve as an input of a 6DoF pose optimization. Given the remaining two sets of correspondences on the source object $\mathbf{O} = {p_1, p_2, ... p_n}$ $\in R^3$ and the target scene $\mathbf{S} = {q_1, q_2, ... q_n}$ $\in R^3$ we then want to find a optimal rigid transformation that aligns the two sets. Specifically, we want to find a rotation $\mathbf{R}$ and a translation $\mathbf{t}$ such that
\begin{equation}
    (\mathbf{R},\mathbf{t}) = \argmin_{\mathbf{R} \in SO(d), t \in \mathbb{R}^3} \sum_{i=1}^n ||(\mathbf{R}p_i + \mathbf{t}) - q_i||^2\,.
\end{equation}

\definecolor{Col1}{RGB}{7, 120, 164}
\definecolor{Col2}{RGB}{29, 206, 212}
\definecolor{Col3}{RGB}{209, 67, 181}
\definecolor{Col4}{RGB}{249, 208, 64}
\definecolor{Col5}{RGB}{237, 146, 48}
\definecolor{Col6}{RGB}{78, 185, 67}
\definecolor{Col7}{RGB}{152, 255, 82}
\definecolor{Col8}{RGB}{128, 234, 133}

\begin{table*}[htbp]
\centering
\caption{Evaluation: keypoint matching of dynamic 3D TSDF patches in \dbName at 95\% recall.}
\label{table:keypoint_matching}
\resizebox{\linewidth}{!}{%
{\renewcommand{\arraystretch}{1.3}%
\begin{tabular}{llccccccccc}
& Method (train) & F1 & Accuracy & Precision & FPR & ER & Top-1 & Top-3 & Top-5 & Top-10 \\
\hline
\arrayrulecolor{white}\hline
\cellcolor{Col3}&\OURS-singlescale 60cm (static) & 71.54 & 62.21 & 57.37 & 70.60 & 75.59 & 2.17 & 4.12 & 5.96 & 17.56\\
\cellcolor{Col4}&\OURS-singlescale 120cm (static) & 74.17 & 66.92 & 60.83 & 61.18 & 66.16 & 3.94 & 4.58 & 8.21 & 20.38\\
\cellcolor{Col5}&\OURS-singlescale 120cm (dynamic) & 78.71 & 74.29 & 67.17 & 46.43 & 51.41 & 6.26 & 7.26 & 9.58 & 27.82\\
\cellcolor{Col6}&\OURS-multiscale (static) & 85.58 & 83.98 & 77.82 & 27.09 & 32.04 & 30.73 & 53.48 & 69.61 & 89.03\\
\cellcolor{Col7}&\OURS-multiscale (dynamic) & \textbf{94.37} & \textbf{94.33} & \textbf{93.61} & \textbf{6.50} & \textbf{11.35} & \textbf{64.10} & \textbf{86.20} & \textbf{93.40} & \textbf{98.30}\\
\\\arrayrulecolor{white}\hline
\end{tabular}}}
\vspace{-0.3cm}
\end{table*}

We solve this optimization using Singular Value Decomposition (SVD). 
The resulting 6DoF transformation gives us a pose that aligns the model to the scene. Qualitative results of our alignment method with corresponding ground truth alignments on some scans of our \dbName-dataset are shown in Figure \ref{fig:results}.

\section{Evaluation}

In the following, we show quantitative experimental results of our method by evaluating it on our newly created \dbName-dataset. 
In the first section, we compare the ability of different methods to match dynamic patches around keypoints on annotated changed objects. Our proposed multi-scale network is then evaluated on the newly-created benchmark for re-localization of object instances.

\begin{figure}[t]
\begin{center}
   \includegraphics[width=1.0\linewidth]{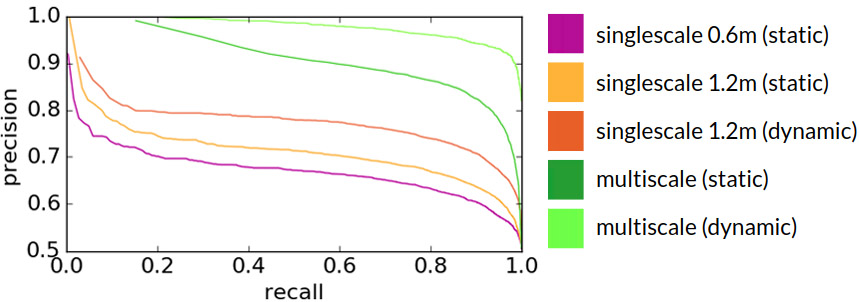}
\end{center}
   \caption{Precision-Recall Curves (PRC) of the dynamic keypoint matching task, corresponding to the different methods evaluated and listed in Table~\ref{table:keypoint_matching}.}
\label{fig:RPC}
\end{figure}

\subsection{Correspondence Matching}

For accurate 6D pose estimation in changing environments, robust correspondence matching is crucial. The feature matching accuracy of different network architectures is reported in Table \ref{table:keypoint_matching}. Each network is pre-trained with static samples (marked as \textit{static}) and then fine-tuned on dynamic patches from the training  (marked as \textit{dynamic}). The F1 score, accuracy, precision, false positive rate (FPR) and error rate (ER) at 95\% recall are listed and visualized with their respective PRC graphs in Figure \ref{fig:RPC}. Additionally to the 1:1 matching accuracy, we also use a top-1 metric: the percentage of top-1 placements of a positive patch given 50 randomly chosen negative patches. Such a metric better represents the real test case of object instance re-localization where several negative samples are compared against a positive keypoint. In can be seen that our multiscale network architecture -- even if only trained with static data -- outperforms all single scale architectures by a large margin and improves further to an $F1$ score of 94.37 if additionally trained with dynamic data.

\begin{table*}[t]
\centering
\caption{Performance of object instance re-localization. Numbers are reported in terms of average \% correct rotation and translation predictions. MTE (Median Translation Error) is measured in meters while the MRE (Median Rotation Error) is in degrees}
\label{table:3D_evaluation}
\resizebox{\linewidth}{!}{
{\renewcommand{\arraystretch}{1.4}%
\begin{tabular}{l|ccc|ccc}
Method (train) & Recall $< $0.1m, \ang{10} & MRE [deg] & MTE [m] & Recall $< $0.2m, \ang{20} & MRE [deg] & MTE [m] \\
\hline
FPFH \cite{Rusu2009} & 2.61 & 7.25 & 0.0645 & 8.36 & 10.57 & 0.0776\\
SHOT \cite{Tombari2010} & 6.79 & 5.35 & 0.0268 & 12.27 & 8.18 & 0.0393\\
3DMatch (dynamic) & 5.48 & 5.81 & 0.0542 & 13.05 & 7.30 & 0.0708\\
\OURS-multiscale (static) & 9.92 & 4.33 & 0.0425 & 17.75 & 6.39 & 0.0545\\
\OURS-multiscale (dynamic) & \textbf{15.14} & 4.75 & 0.0437 & \textbf{23.76} & 6.08 & 0.0547 \\
\end{tabular}}}
\end{table*}

\begin{table}[htbp]
\centering
\caption{Matching accuracy of the different methods for different instance categories at $< $0.2m, \ang{20} and our method trained on static (\OURS-S) and dynamic data (\OURS-D). See supplementary for detailed class description.}
\label{table:3D_cateogries}
\resizebox{\linewidth}{!}{%
{\renewcommand{\arraystretch}{1.4}%
\begin{tabular}{lccccc}
class & FPFH & SHOT & 3DMatch & \OURS-S & \OURS-D\\
\hline
seating & 5.08&  12.71&  6.78&   14.41&   21.19\\
table & 9.33&   5.33&   21.33&    25.33&   29.33\\ 
items & 5.06&    13.92&  7.59&   11.39&   16.46\\
bed / sofa &   56.52&  21.74&    34.78&  34.78&  47.83\\
cushion & 0.00&   15.52&  8.62&    8.62&  10.34\\
appliances & 11.11&    16.67 &   33.33&  44.44&  55.56\\
structure & 0.00&    0.00&  8.33&  16.67&  33.33\\
\hline
\textbf{avg.} &12.44& 12.27& 17.25&  22.23 & \textbf{30.58}\\
\end{tabular}}}
\end{table}

\subsection{Object Instance Re-localization}

In the following we discuss results on our newly created benchmark that has been carried out on the test set of \dbName{ } which is provided with the data. We evaluate our method against hand-crafted features from PCL \cite{Rusu2011} such as SHOT \cite{Tombari2010} and FPFH \cite{Rusu2009,Aldoma2011}. A transformation of each object instance is computed separately by (1) sampling keypoints, (2) extracting descriptors at each keypoint followed by a (3) correspondence matching and (4) RANSAC-based filtering. A learned baseline we evaluate against is 3DMatch \cite{Zeng2016}. It computes a feature given a patch around a keypoint. We trained 3DMatch on $30\times30\times30$ static positive and negative patches of $\SI{30}{\cm}$ size generated with our dataset as described in the original paper. 
We evaluate the predicted rotation $\mathbf{R_p}$ and translation $\mathbf{t_p}$ against the ground truth annotation $\mathbf{R_{GT}}$ and $\mathbf{t_{GT}}$ according to equation \ref{eq:rotation_evaluation} and \ref{eq:translation_evaluation}. An instance has successfully been aligned if the alignment error for the translation $\mathbf{t_{\Delta}}$ and rotation $\mathbf{R_{\Delta}}$ are lower than $t \leq \SI{10}{\cm}$, $r \leq \SI{10}{\degree}$ or $t \leq \SI{20}{\cm}$, $r \leq \SI{20}{\degree}$. Please note that respective symmetry are considered in the error computation: 

\begin{equation}
\label{eq:translation_evaluation}
\mathbf{t_{\Delta}} = \mathbf{t_p} - \mathbf{t_{GT}}
\end{equation}

\begin{equation}
\label{eq:rotation_evaluation}
\mathbf{R_{\Delta}} = \mathbf{R_{p}^{-1}}\mathbf{R_{GT}} \rightarrow \text{axis angle}
\end{equation}

Evaluation results for all object instances are listed in Table \ref{table:3D_evaluation} and Table \ref{table:3D_cateogries}. While classical hand-crafted methods still perform reasonable well -- especially for more descriptive objects such as sofas and beds -- our method outperforms them with a large margin. Qualitative results are shown in Figure \ref{fig:results}.

\section{Conclusion}

In this work, we release the first large-scale dataset of real-world sequences with temporal discontinuity that consists of multiple scans of the same environment. 
We believe that the new task of object instance re-localization (\OURS) in changing indoor environments is a very challenging and particularly important task, yet to be further explored. 
Besides 6D object instance alignments in those changing environments, \dbName{ } comes with a large variety of annotations designed for multiple benchmark tasks including -- but not limited to -- persistent dense and sparse SLAM, change detection or camera re-localization. 
We believe that \dbName{ } helps the development and evaluation of these new algorithms and we are excited to see more work in this domain to, in the end, accomplish persistent, long-term understanding of indoor environments.

\section*{Acknowledgment}
We would like to thank the volunteers who helped with 3D scanning, all expert annotators, as well as J{\"u}rgen Sturm, Tom Funkhouser and Maciej Halber for fruitful discussions.
This work was funded by the Bavarian State Ministry of Education, Science and the Arts in the framework of the Centre Digitisation.Bavaria (ZD.B), the ERC Starting Grant Scan2CAD (804724), TUM-IAS for the Rudolf M{\"o}{\ss}bauer Fellowship, and a Google Research and Faculty award.

% 8 pages without references, uncomment \verb'\cvprfinalcopy', use \eg and \etal (no double periods)

{\small
\bibliographystyle{ieee_fullname}
\bibliography{egbib}
}

\clearpage

\section{Supplemental Material}

In this supplemental document, we provide additional information about the proposed dataset such as statistics, scene examples and a detailed description about the annotation process. 

\section*{Dataset}

\paragraph{Scanning Interface}

\begin{figure}[htbp]
\begin{center}
   \includegraphics[width=1.0\linewidth]{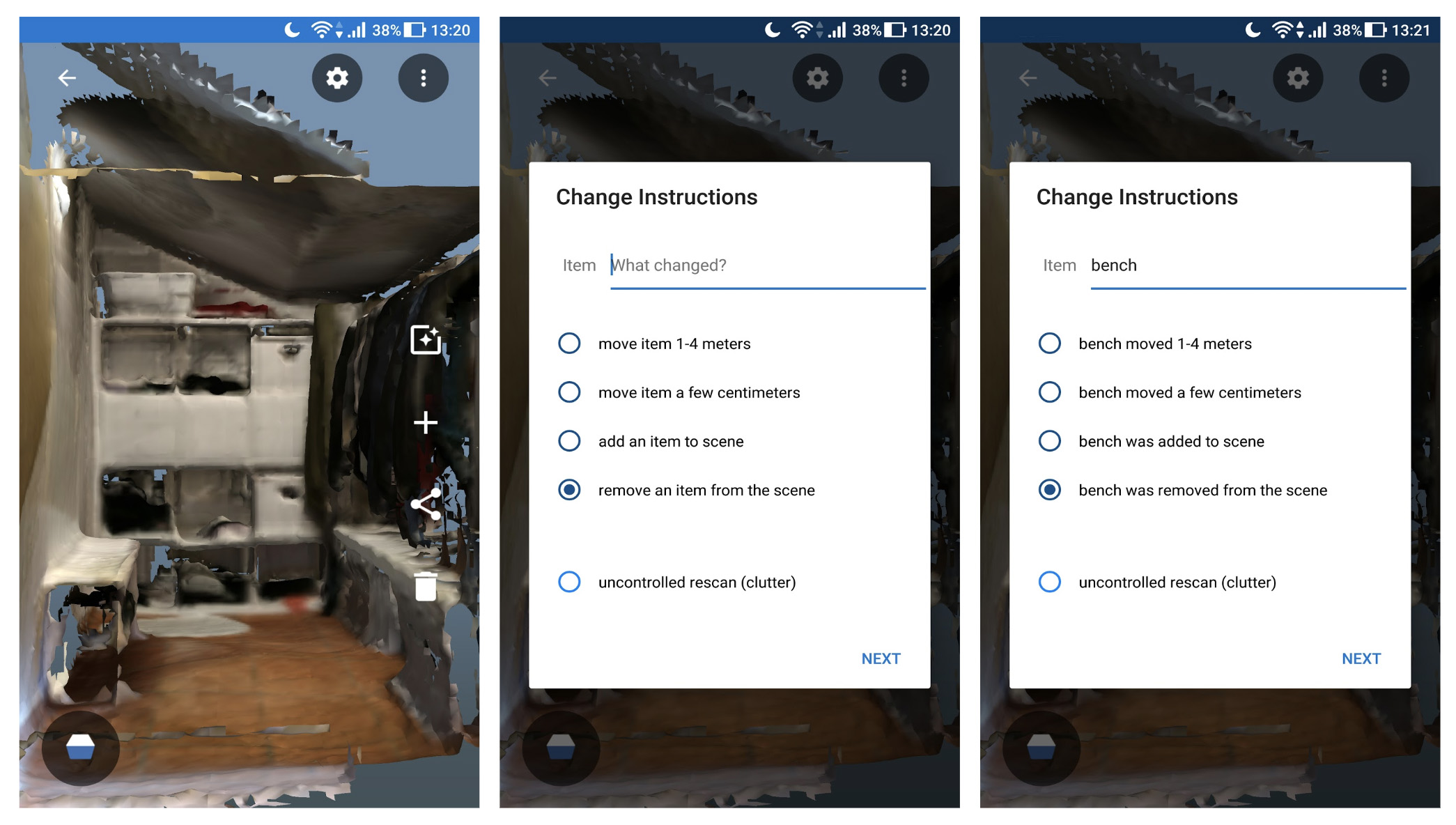}
\end{center}
   \caption{Annotation Interface used for data acquisition.}
\label{fig:constructor}
\end{figure}

We tailored a mobile app running on Google Tango with pre-annotation functionality as a scanning interface (see Figure \ref{fig:constructor}). Some users gave lightweight instructions on the scene changes; these instructions served as guidelines later in the annotation process.

\paragraph{Scene Matching and Alignment}

For each uploaded scan, scene candidates are computed. Since a 3D scene matching is expensive scan pairs are found in 2D instead by conducing a similarity search in the texture uv-map of the mesh. 
These matches are then to be manually adjusted. Once the reference for each scene is assigned, the IMU normalized scans are globally registered via a coarse to fine correspondence-based 2D ICP together with RANSAC and refined with a global 3D ICP. An additional verification as well as an optional manual, keypoint based alignment ensures high quality.

\paragraph{Preprocessing}

\begin{figure}[htbp]
\begin{center}
   \includegraphics[width=1.0\linewidth]{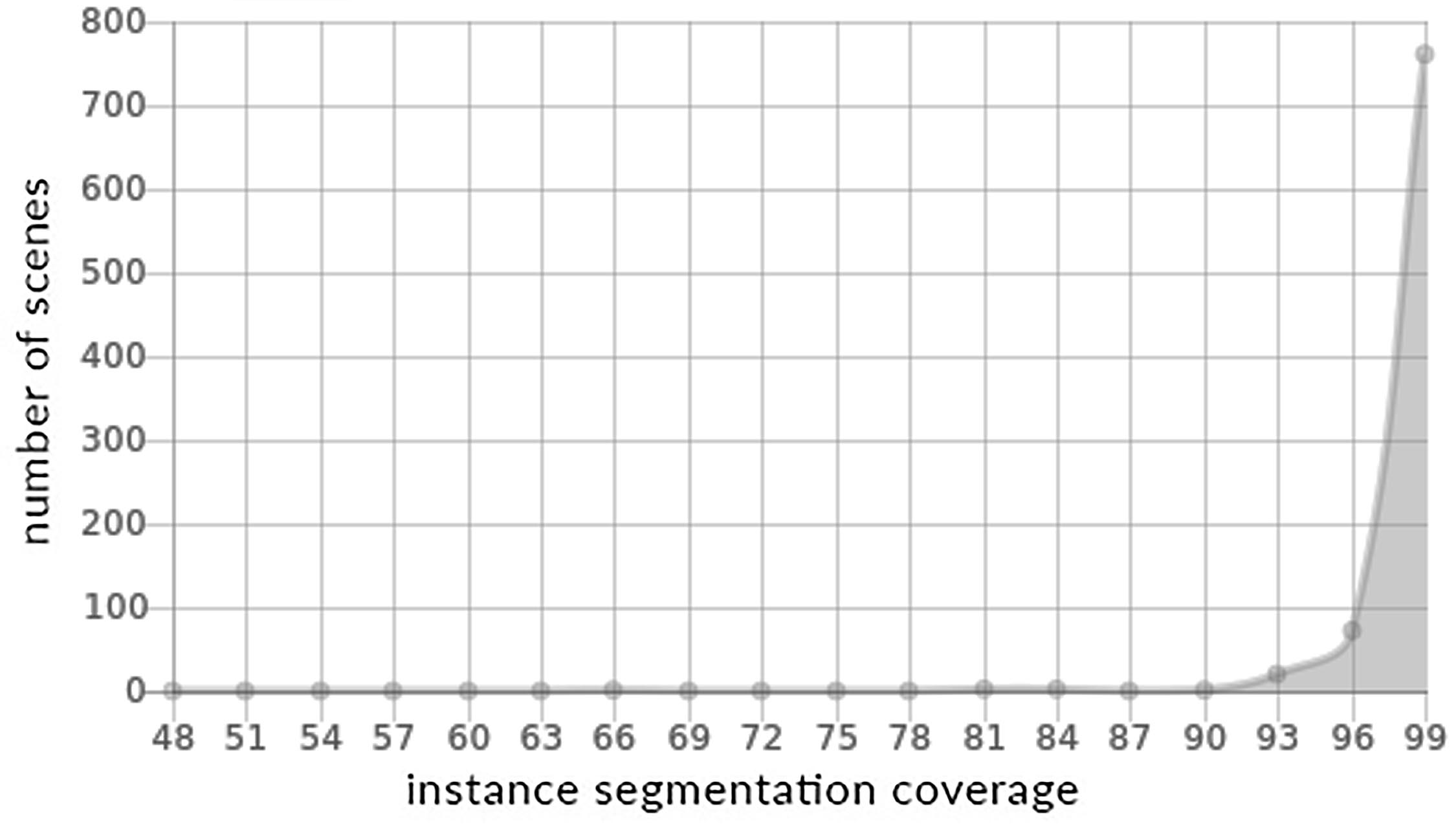}
\end{center}
   \caption{Number of scans with corresponding instance annotation coverage.}
\label{fig:annotation_coverage}
\end{figure}

\begin{figure*}[htbp]
\begin{center}
   \includegraphics[width=0.85\linewidth]{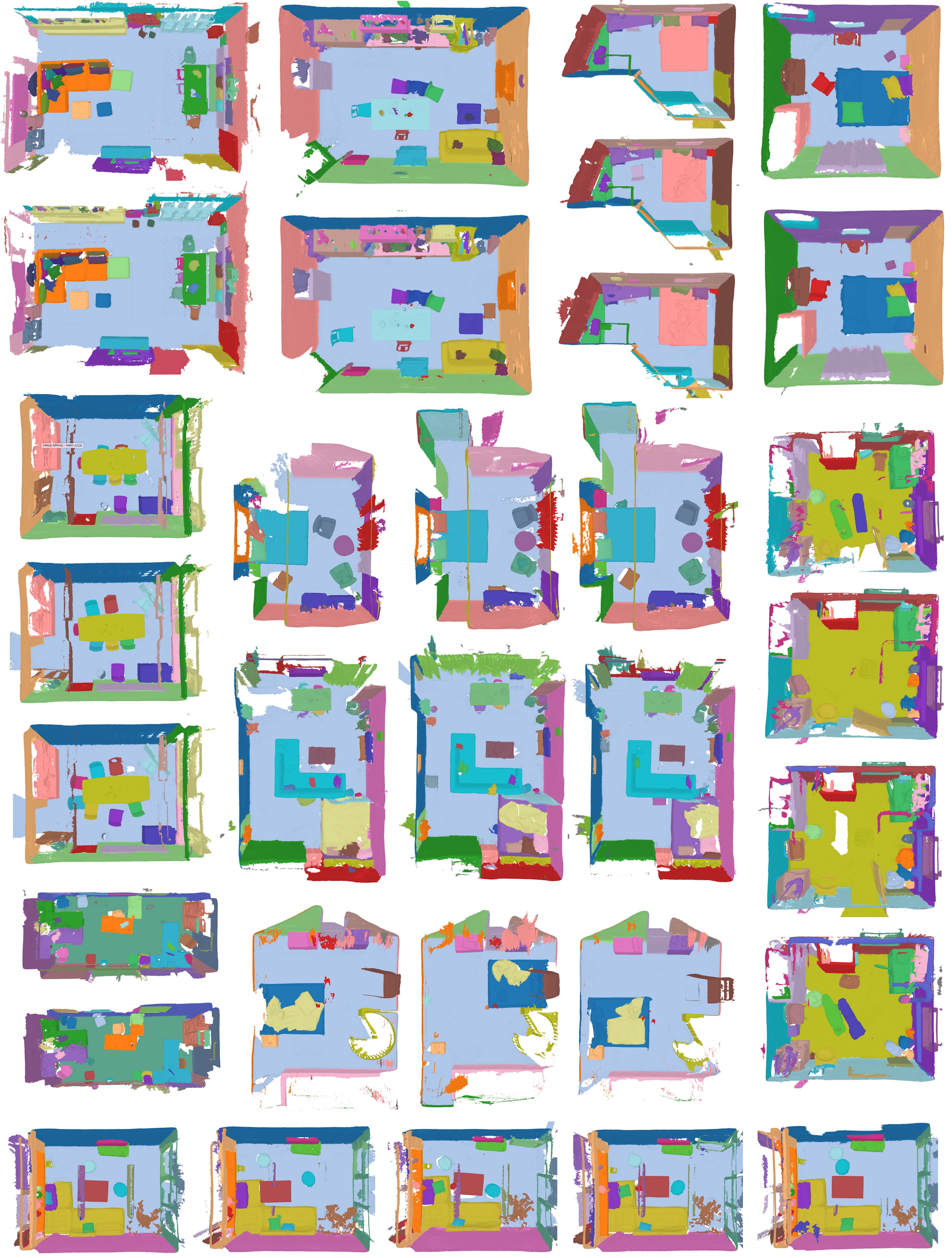}
\end{center}
   \caption{Visualization of different annotated instances in scans of \dbName.}
\label{fig:annotation_suppl}
\end{figure*}

Additionally to a server-side offline processing of the RGB-D sequences that results in texture mapped 3D reconstructions, an offline 3D segmentation is triggered. This 3D segmentation is utilized by the semantic segmentation interface proposed by Dai \etal \cite{Dai2017}. Further, this 3D segmentation -- together with the aforementioned 3D alignment -- serves as the basis for the propagation of semantic labels from the references to the re-scans and after a manual clean-up procedure results in the final instance segmentation shown in Figure \ref{fig:annotation_suppl}. In the current snapshot of the dataset almost all of our scans have an instance segmentation coverage of above 90\% (see Figure \ref{fig:annotation_coverage}) with an average scene coverage $>$ 98\%. In total, \instanceCount instances are annotated with 534 unique labels.

\paragraph{RGB-D sequences}

\begin{figure*}[htbp]
\begin{center}
   \includegraphics[width=\linewidth]{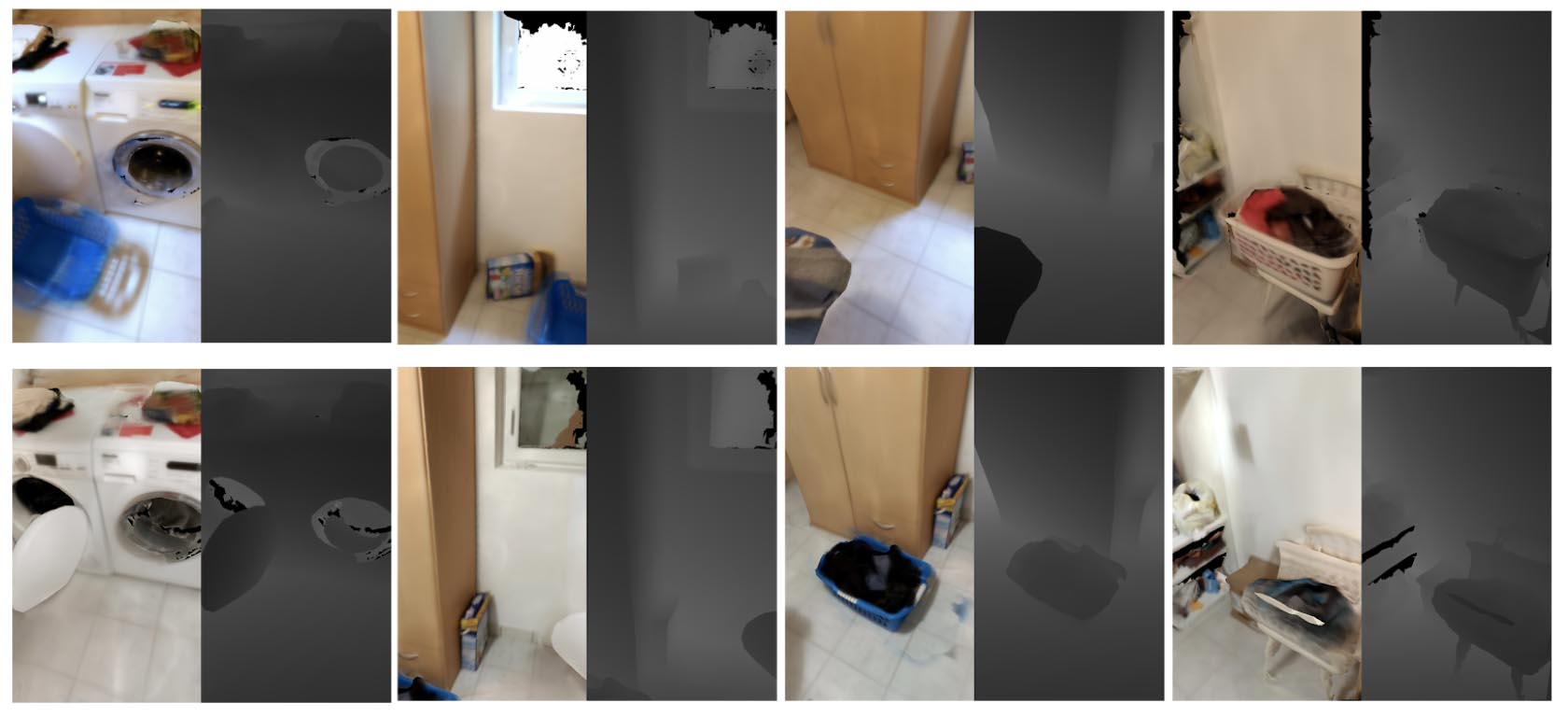}
\end{center}
   \caption{Example 2D projections of the color and depth of two corresponding reconstructions with natural scene changes.}
\label{fig:projections2D}
\end{figure*}

\begin{figure*}[htbp]
\begin{center}
   \includegraphics[width=0.9\linewidth]{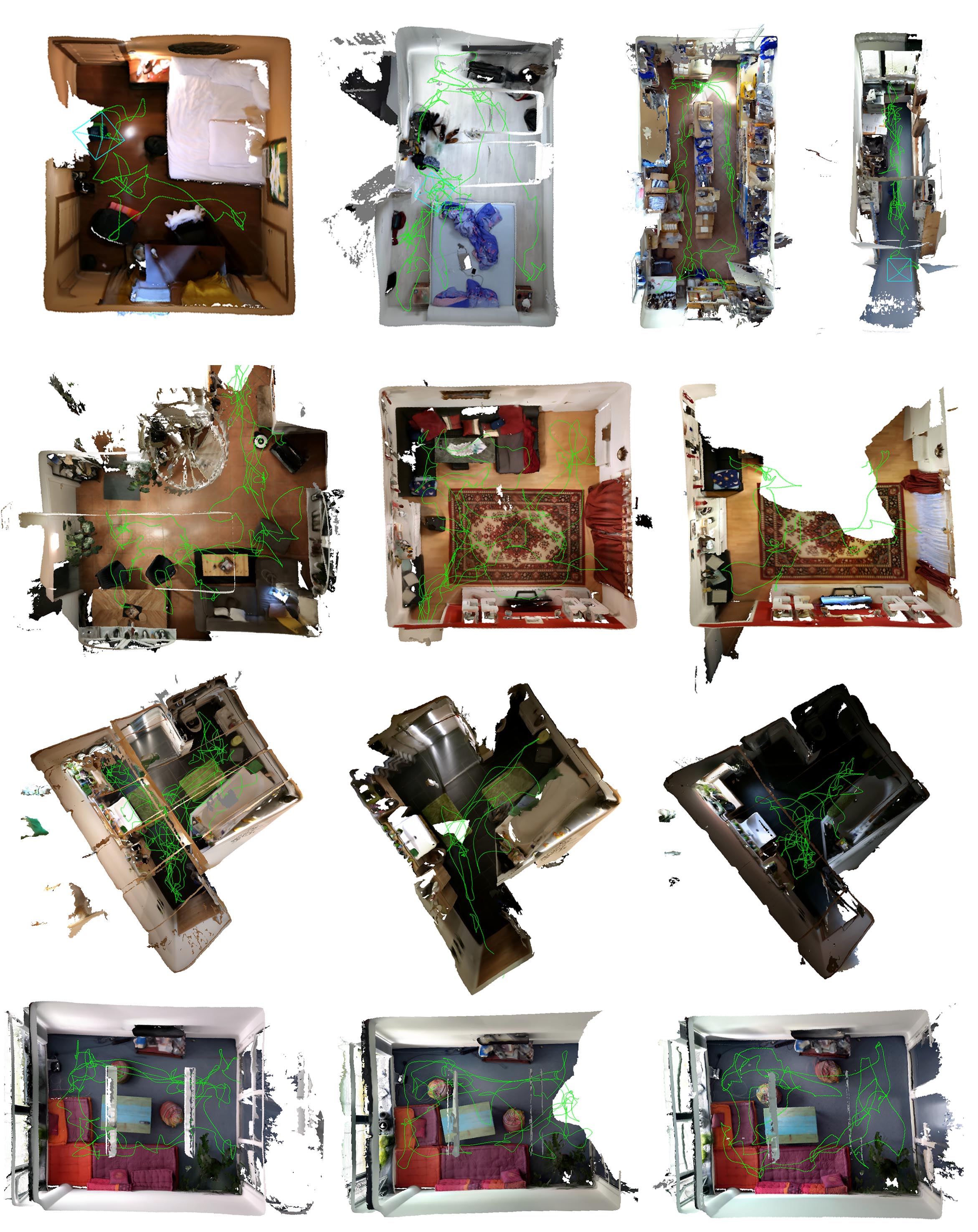}
\end{center}
   \caption{SLAM: Different 3D Scenes with Camera Trajectories in green used for training and generation of the static TSDF samples.}
\label{fig:SLAM}
\end{figure*}

Our dataset consists of around \frameCount calibrated RGB-D and depth images. Since raw RGB and depth sequences from Tango are of varying frame rates and spatial resolution, a spatial and temporal calibration procedure of the raw images is applied. Further, to remove rectification lines present in Google Tango depth images, a median filter is used before the spatial calibration of the images. Camera trajectories for SLAM are visualized in Figure \ref{fig:SLAM} and since global scene-to-scene mappings are provided these can easily be transferred into the same coordinate systems (see last two rows of Figure \ref{fig:SLAM}). Further, we also show 2D projections of our textured 3D models with aligned camera poses in Figure \ref{fig:projections2D}.

\paragraph{Scene Type}

Further, instead of assigning a $1-n-$relationship of different room types per scene, our 3D reconstructions are annotated with $m$ corresponding scene functionalities (sleeping, eating, working, etc.) in an $n-m$ fashion. This shows the high variety of scenes in \dbName.

\paragraph{Instance Change Annotation}

\begin{figure*}[htbp]
\begin{center}
   \includegraphics[width=0.92\textwidth]{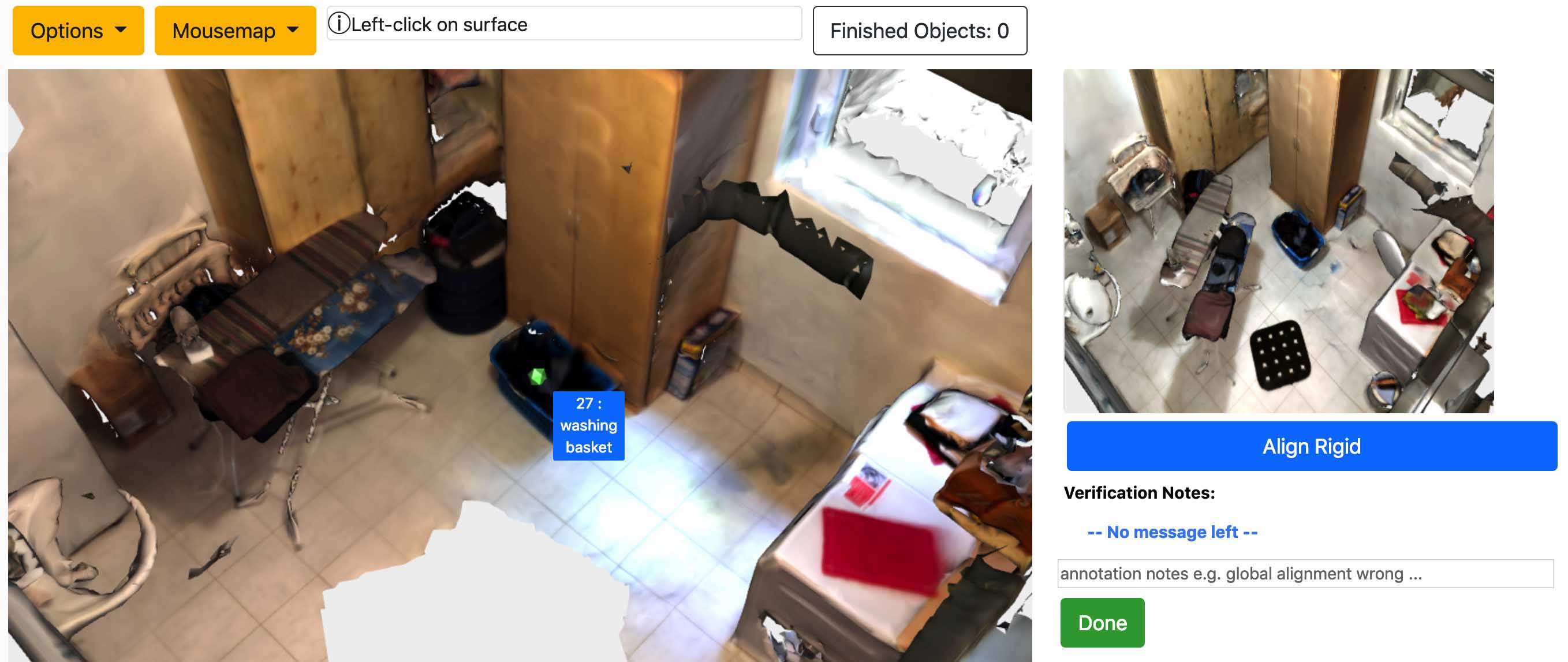}
\end{center}
   \caption{Instance Change Annotation Tool: Overview and selection view of the instance alignment annotation.}
\label{fig:annotation_tool_3D}
\end{figure*}

The annotation interface for annotating instance changes is a web-based tool (Figure \ref{fig:annotation_tool_3D}) where each scene is rendered next to its corresponding reference. When an object is selected in the re-scan (see green dot) its instance segmentation from the reference scan is automatically segmented. Please note, that this requires the instance IDs to be consistent across scans of the same environments. Hovering over the objects gives shows the label and the ID of the instance and allows to potentially fix the underlying semantic segmentation. In the alignment view this instance is then shown next to the re-scan such that corresponding keypoints can easily be selected. Once enough keypoints are set, a Procrustes based alignment (Kabsch algorithm) is triggered that computes a transformation that aligns the object to the scene. For non-rigid changes and removed as well as added objects the instances IDs are tracked.

\paragraph{Symmetry}

A subset of the changed objects in the dataset are symmetric. We follow the symmetry annotation described in Avetisyan \etal \cite{Avetisyan2018} and categorize each object's rotational symmetry around the canonical axis to the classes $C_2$, $C_4$ and $C_{\infty}$. 22\% of the objects have a symmetry as listed in Table \ref{table:symmetry_table}. We take this into account when evaluating the predictions against ground truth poses.

\begin{table}[htbp]
\centering
\caption{Symmetry properties of the instances in \dbName}
\label{table:symmetry_table}
{\renewcommand{\arraystretch}{1.3}%
\begin{tabular}{l|llll}
symmetry & none & $C_2$ & $C_4$ & $C_\infty$ \\
$\#$ scans & 1513  & 220 & 82 & 132\\
\end{tabular}}
\end{table}

\paragraph{Statistics}

\begin{figure}[htbp]
\begin{center}
   \includegraphics[width=1.0\linewidth]{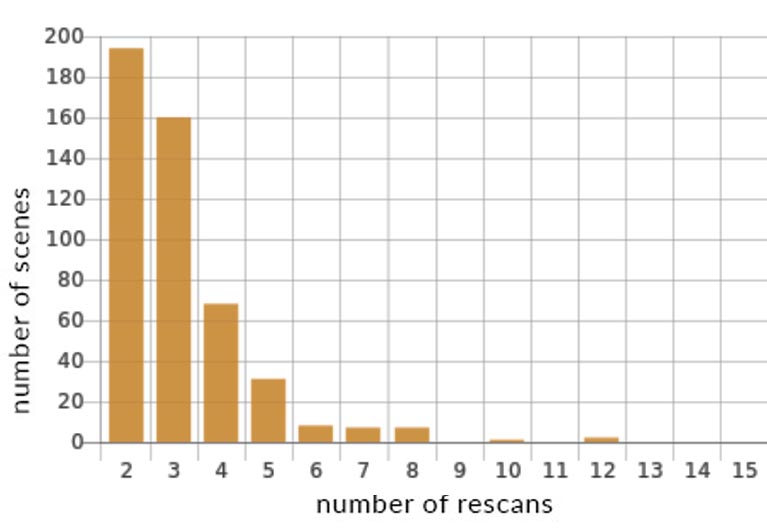}
\end{center}
   \caption{Number of scenes vs. number of re-scans.}
\label{fig:rescans_stats}
\end{figure}

A focus during data acquisition was the capturing of a variety of realistic scene changes in controlled and uncontrolled environments over a time span of more than $12$ months. The number of references scenes with re-scanning frequencies are plotted in Figure \ref{fig:rescans_stats}. Further, \objectCount instance transformations -- of \objectInstanceCount different objects -- are provided with the data. But since the transformations give the object pose from the reference to one re-scan the alignment for another re-scan can easily be computed. For evaluation these changed object categories are mapped to 9 different classes as listed in \ref{table:instance_cateogries}.

\begin{table}[htbp]
\centering
\caption{Description of instance mapping used in the evaluation}
\label{table:instance_cateogries}
{\renewcommand{\arraystretch}{1.3}%
\begin{tabular}{lccccc}
class & description\\
\hline
seating & different chairs, stools, benches\\
table / cabinet & different tables, commode, shelves\\
bed / sofa & upholstery, sofas, beds\\
appliances & appliances, sanitary equipment\\
cushions & pillows, bean bags, ottoman\\
items & small and portable items, boxes\\
structure & windows, doors\\
\end{tabular}}
\end{table}

\begin{figure}[htbp]
\begin{center}
   \includegraphics[width=1.0\linewidth]{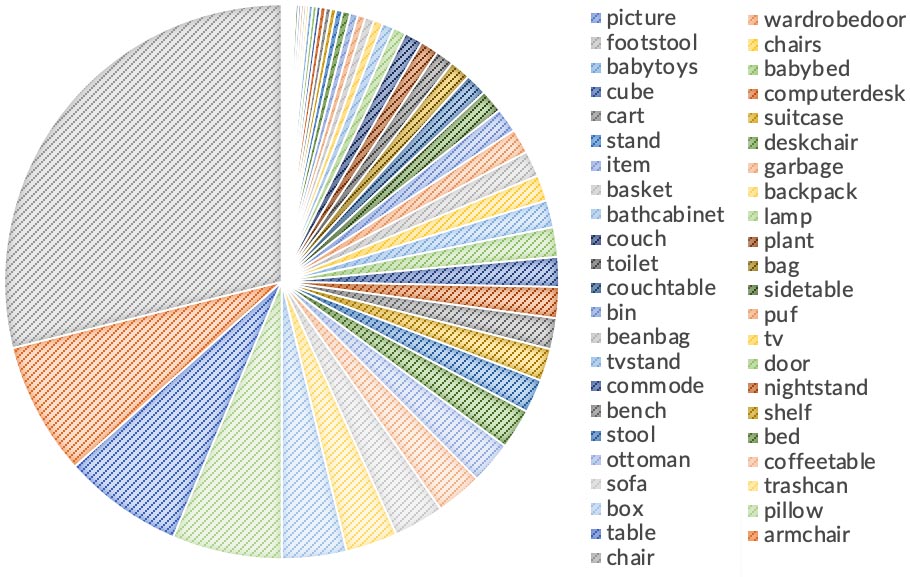}
\end{center}
   \caption{Object statistics.}
\label{fig:obj_stats}
\end{figure}

\begin{figure}[htbp]
\begin{center}
   \includegraphics[width=1.0\linewidth]{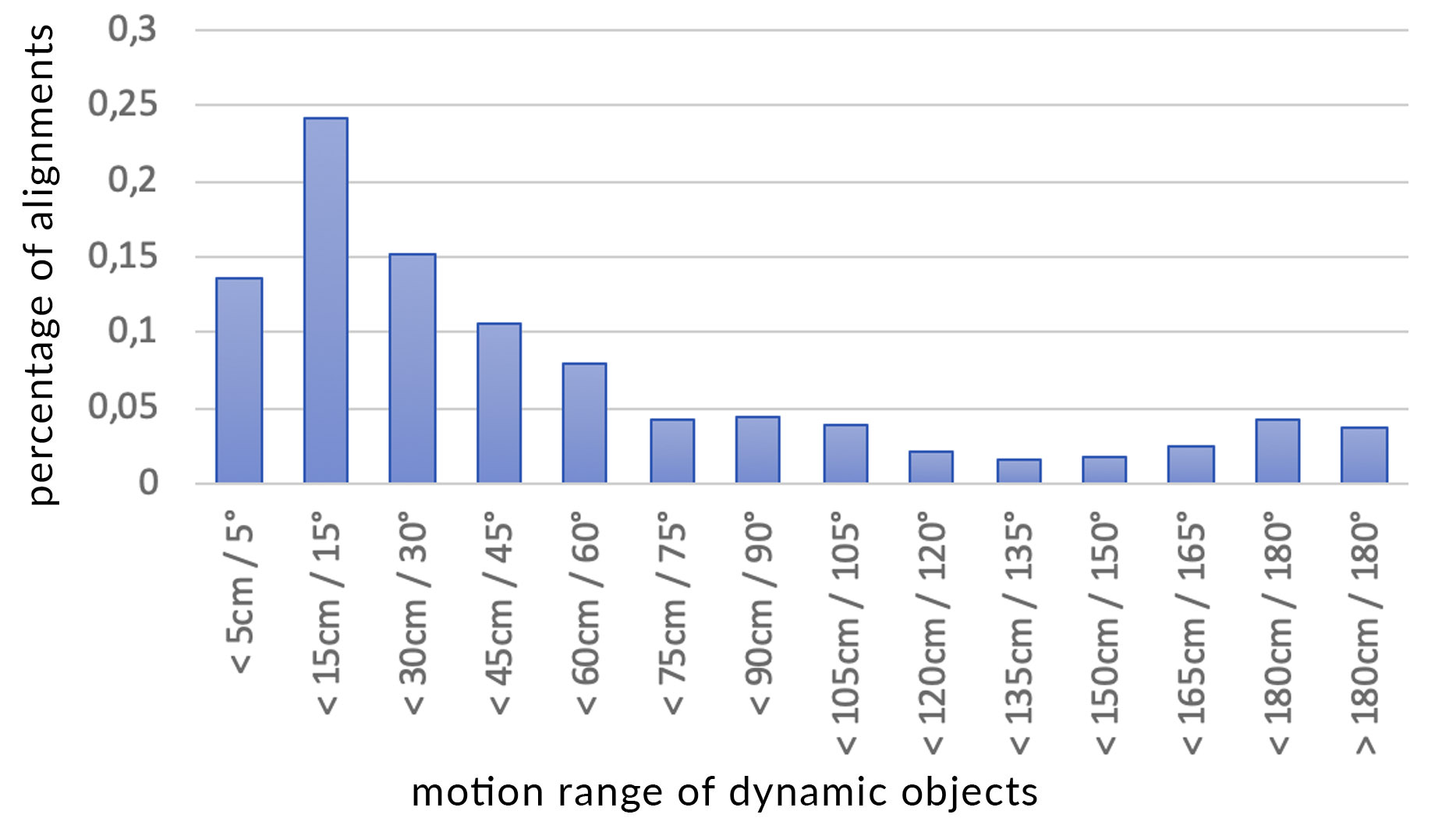}
\end{center}
   \caption{Histogram of object instance alignment binned using respective transformation or rotation change.}
\label{fig:motion_stats}
\end{figure}

 These changed object instances are labelled with \objectCateogryCount different categories. The majority of instances include movements of objects and more portable furniture items such as chairs, pillows, boxes or smaller tables. Naturally, these objects involve most human interaction. Figure \ref{fig:motion_stats} gives an overview of the motion of these annotated objects. However, we also annotated objects that slightly change their appearance over time such as toilets. Detailed statistics are given in Figure \ref{fig:obj_stats}.

\begin{figure*}[htbp]
\begin{center}
\includegraphics[width=0.95\linewidth]{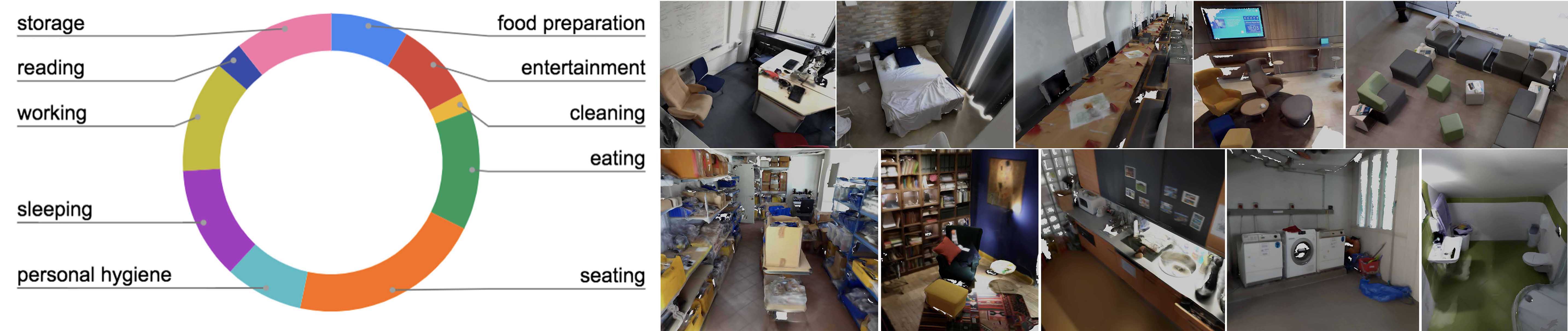}
\end{center}
\caption{List of mutually inclusive scene functionalities with corresponding visual examples, from top left to bottom right: (a) working, (b) sleeping, (c) eating, (d)  entertainment, (e) seating, (f) storage, (g) reading, (h) food preparation, (i) cleaning and (j) personal hygiene.}
\label{table:functionalities}
\end{figure*}

\end{document}